%% file: main.tex
\documentclass[11pt,a4paper]{article}
\usepackage[hyperref]{emnlp2020}
\usepackage{times}
\usepackage{latexsym}

\usepackage{algorithm}%
\usepackage{algpseudocode}%

\usepackage{microtype}
\usepackage{booktabs}
\usepackage{courier}
\usepackage{linguex}
\usepackage{color}
\usepackage{listings}
\usepackage{array}
\usepackage{multirow}
\usepackage{graphicx}
\usepackage{todonotes}
    
\aclfinalcopy %
\def\aclpaperid{26} %

\newcommand{\abb}[1]{\textbf{\texttt{#1}}}
\newcommand\cometemoji{\raisebox{-2pt}{\includegraphics[width=0.9em]{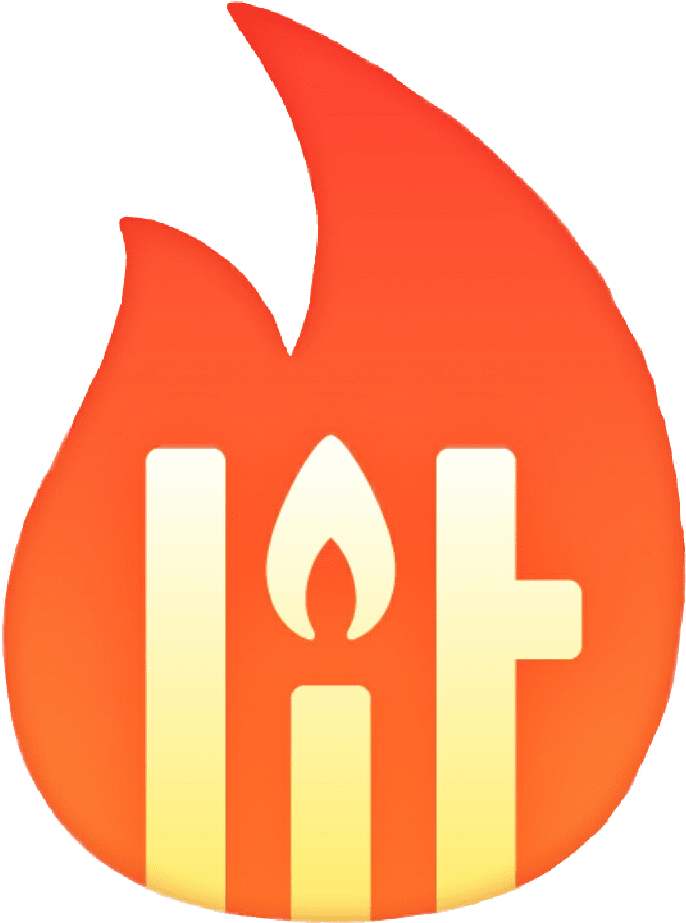}}}

\title{Linguistically-Informed Transformations (LIT\cometemoji{}): A Method for Automatically Generating Contrast Sets}

\author{$\{$Chuanrong Li$^{\heartsuit}$, Lin Shengshuo$^{\heartsuit}$, Leo Z. Liu$^{\clubsuit}$, Xinyi Wu$^{\heartsuit}$, Xuhui Zhou$^{\heartsuit}$ $\}$\thanks{~~Equal contribution from bracket authors, sorted by author's last name}, \\ \textbf{Shane Steinert-Threlkeld}$^{\heartsuit}$\\
\\
  $^{\heartsuit}$Department of Linguistics, University of Washington \\
  $^{\clubsuit}$Paul G.\ Allen School of Computer Science \& Engineering, University of Washington \\
 {\tt \{licor,shuo2019,xywu,xuhuizh,shanest\}@uw.edu},\\ 
 \tt{zeyuliu2@cs.washington.edu }}

\begin{document}
\maketitle
\begin{abstract}

Although large-scale pretrained language models, such as BERT and RoBERTa, have achieved superhuman performance on in-distribution test sets, their performance suffers on out-of-distribution test sets (e.g., on contrast sets). Building contrast sets often requires human-expert annotation, which is expensive and hard to create on a large scale. In this work, we propose a Linguistically-Informed Transformation (LIT) method to automatically generate contrast sets, which enables practitioners to explore linguistic phenomena of interests as well as compose different phenomena. Experimenting with our method on SNLI and MNLI shows that current pretrained language models, although being claimed to contain sufficient linguistic knowledge, struggle on our automatically generated contrast sets. Furthermore, we improve models' performance on the contrast sets by applying LIT to augment the training data, without 
affecting performance on the original data.\footnote{See code in \url{https://github.com/leo-liuzy/LIT_auto-gen-contrast-set}}

\end{abstract}

\section{Introduction}

\input{sec_introduction}

\section{Related Work}
\input{sec_related_work}

\section{Generating Contrast Sets}
\input{figtabs/fig_LIT_pipeline}
\input{sec_method}

\section{Experiments}
\input{sec_experiment}

\section{Discussion and Analysis}
\input{sec_discussion_and_analysis}

\section{Conclusion}
\input{sec_conclusion}

\section*{Acknowledgments}

We appreciate useful feedback from Noah A. Smith, Luke Zettlemoyer, Jungo Kasai, Yizhong Wang, and anonymous reviewers.

\bibliography{anthology,emnlp2020,myrefs}
\bibliographystyle{acl_natbib}

\newpage

\appendix

\section{Transformations in LIT}
\label{app:phenomenon}
In Table~\ref{tab:rule_list}, we show the full list of transformation that LIT can generate. This is not the full capability of LIT, and more transformations are possible as long as the linguistic phenomena are allowed by the ERG grammar.
\input{appendices/tab_rule_list}

\section{Annotator Agreement}
\label{app:agreement}
To confirm the quality of the generated sentences, we recruit experienced graduate students as our annotators. For each phenomenon, we randomly select 50 sentences and have three annotators to judge. Given a phenomenon, each annotator is asked to judge whether they deem the generated (and selected) sentence as grammatical. The gold labels (i.e., grammatical or not) are determined by majority vote. For the it-cleft phenomenon, LIT can generate sentences that emphasize the first theta argument (ARG1) or the second theta argument (ARG2) of the verb for the main clause. Annotation results are shown in the Table~\ref{tab:annotation}.
\input{appendices/tab_annotation}

\section{Full experiments results}
In this section, we show the detailed evaluation results from all models --- {\small\verb|bert-base-uncased|}, {\small\verb|bert-large-uncased|}, {\small\verb|roberta-base|}, and  {\small\verb|roberta-large|} --- trained seperately on two scenarios --- ORI and AUG.
\label{app:full results}
\input{appendices/mnli_tables}
\input{appendices/snli_tables}

\end{document}

%% file: sec_introduction.tex
Large-scale pretrained language models have given remarkable improvements to a wide range of NLP tasks \citep{Peters:2018,howard-ruder-2018-universal, Devlin2019BERTPO,Liu2019RoBERTaAR,radford2019language}. However, the results are questionable, since those models take advantage of lexical cues (and other heuristics) in the datasets, which can make them right for wrong reasons \citep{gururangan-etal-2018-annotation,mccoy-etal-2019-right}. Therefore, the concept of evaluating models on contrast sets \citep{Gardner2020EvaluatingNM} and the creation of generalization tests \citep{Kaushik2020Learning} is critical for building a robust NLP system. Those test sets are usually manually created, which requires significant human effort, and so is hard to do on a large scale.

\input{figtabs/fig_breaking_bert}

In this work, we propose Linguistically-Informed Transformations (LIT) to create contrast sets automatically.  Our method can perturb the original examples and generate various types of contrastive examples, with a wide choice of linguistic phenomena. Furthermore, our tool supports compositional generalization tests. Namely, researchers can choose transformations from a set of basic linguistic phenomena and modify original sentences with an arbitrary combination of those basic transformations.

To demonstrate the utility of LIT, we focus on the natural language inference (NLI) task, a central task to many NLP applications. We apply LIT to generate contrast sets for SNLI \citep{bowman-etal-2015-large} and MNLI \citep{N18-1101} using seven linguistic phenomena. Human experts' rating show that our generated data is high-quality for basic transformations and for most of the compositional transformations. See Appendix~\ref{app:agreement} for more details. 

With our generated contrast sets, we show that pretrained language models, despite having `seen' huge quantities of raw text data, fail on simple linguistic perturbations. As shown with an example in Figure~\ref{fig:breaking_bert}, `decoupling' tenses of the premise and hypothesis breaks BERT's prediction. Our analysis not only shows the inadequate coverage of SNLI and MNLI datasets but also reveals the deficiency of current pretraining-and-finetuning paradigms. Compared to previous work showing that BERT is not robust and fails to generalize on out-of-distribution test sets \citep{mccoy-etal-2019-right,zhou2019evaluating, Jin2019IsBR}, our method provides a more fine-grained picture showing on which phenomenon the models fail. In summary, our contributions are:

\begin{itemize}
    \item We provide a method for automatically generating phenomenon-specific contrast sets, which helps NLP practitioners better understand pre-trained language models.
    
    \item We further apply LIT to augment SNLI and MNLI training data, which improves models' performance on out-of-distribution test sets without sacrificing the models' performance on the in-distribution test set.
    
    \item We demonstrate that, in the current pretraining paradigm, traditional linguistic methods are valuable for their ability to measure and promote robustness and consistency in data-driven models. %
    
\end{itemize}

After discussing several areas of related work in Section~\ref{related}, we describe LIT in step-by-step detail (Section~\ref{lit-method}).  We then apply LIT to SNLI and MNLI (\ref{applying-lit}) before evaluating BERT and RoBERTa on both simple (\ref{simple-transformations}) and compositional (\ref{compositional-transformations}) transformations.  We conclude (Section~\ref{discussion}) by discussing limitations of LIT and future directions.

%% file: figtabs/fig_breaking_bert.tex
\begin{figure}[t]
    \centering
    \includegraphics[width=\linewidth]{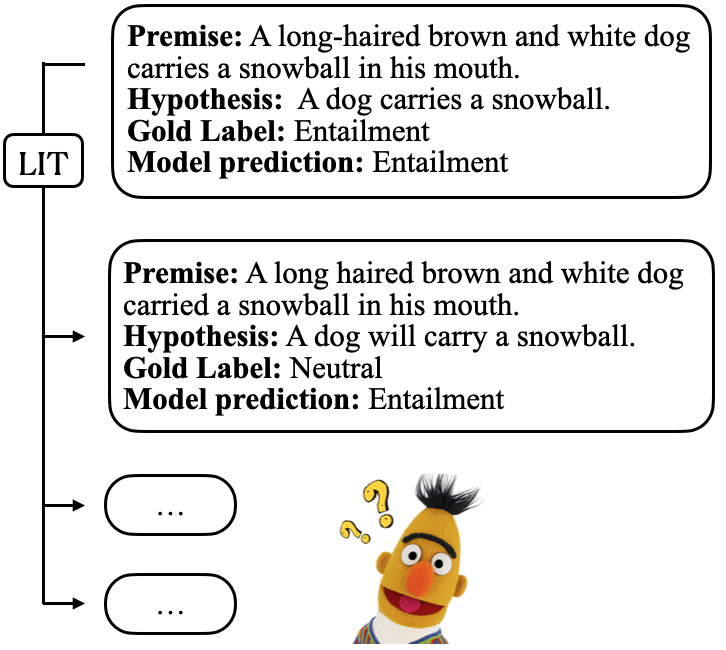}
    \caption{Example of BERT making wrong prediction on LIT-transformed data but correct prediction on the original datum.
    The detailed transformed datum includes a premise modified to past tense and a hypothesis with future tense.
    The true label correspondingly changes to \textit{neutral}.
    LIT also generates multiple transformation results at once for a single original datum; we include only one detailed example here for simplicity of the illustration.}
    \label{fig:breaking_bert}
\end{figure}

%% file: sec_related_work.tex
\label{related}

\paragraph{NLI Model Diagnosis} Our work builds on works diagnosing and improving NLI models with automatically augmented instances \citep{mccoy-etal-2019-right, min2020augmentation}. While most of these works apply simple methods such as templates to generate new instances, which limits the phenomena covered, our method has a wider coverage and can be easily extended. 

\paragraph{Contrast Sets} Contrast sets \citep{Gardner2020EvaluatingNM} serve to evaluate a models' true capabilities by evaluating on out-of-distribution data since previous in-distribution test sets often have systematic gaps, which inflate models' performance on a task \cite{gururangan-etal-2018-annotation, geva-etal-2019-modeling}. The idea of contrast sets is to modify a test instance to a minimum degree while preserving the original instance's syntactic/semantic artifacts and changing the label. Typically, the authors of the dataset create the contrast set manually. We show that a precision grammar, namely ERG \citep{copestake-flickinger-2000-open}, can be used to automate this process while preserving the authors' benefit of choosing the perturbations of interest. 

\paragraph{Adversarial Datasets} Another line of work addressing the problem of current models' super-human performance on in-distribution test sets focuses on adversarial methods. \citet{bras2020adversarial} uses an adversarial filtering algorithm to reduce spurious bias in the dataset to avoid models relying on such patterns. \citet{Dinan_2019} shows that a human-in-the-loop adversarial training framework significantly improves models' robustness. And \citet{jin2019bert} shows that current pretrained language models are not robust under simple lexical manipulations. Adversarial methods generate test instances automatically, which can be applied to augment the training data \citep{jin2019bert, Dinan_2019}. However, these adversarial methods introduce specific models in the loop, which might also bias the test set.

%% file: figtabs/fig_LIT_pipeline.tex
\begin{figure*}[t]
    \centering
    \includegraphics[width=\linewidth]{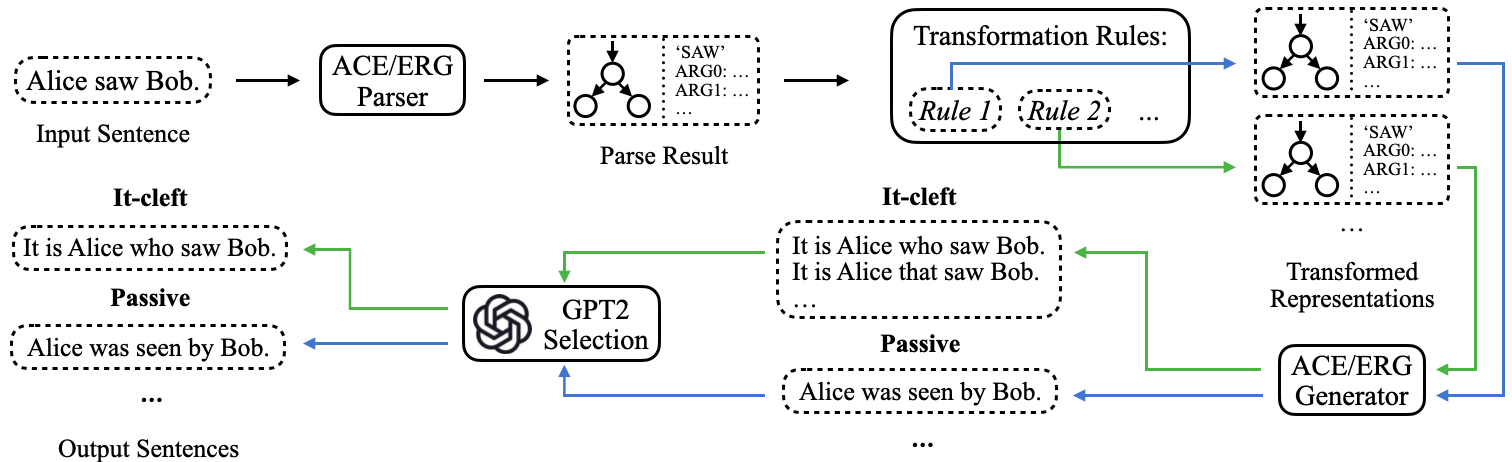}
    \caption{General pipeline of LIT system exemplified with one input sentence.
    The parse result includes both syntax and semantics.
    The transformation rules produce one transformed representation per phenomenon.
    A set of sentences, all grammatical according to ERG, is generated for each transformed representation.
    One sentence per phenomenon is selected as the final output sentence.
    We include two ``\textit{Rule}"s for illustration purpose; LIT includes more transformation rules and can be extended for more phenomena.}
    \label{fig:LIT_pipeline}
\end{figure*}

%% file: sec_method.tex
\label{lit-method}

We propose a new Linguistically-Informed Transformation (LIT) method for large-scale automatic generation of contrast sets.
LIT 1) parses the input sentence for both syntax and semantics, 2) produces transformed syntax and semantics for each linguistic phenomenon, 3) generates perturbed sentences corresponding to the transformed syntax/semantics, 4) and selects the best surface sentence for each phenomenon. The full pipeline is shown in Figure~\ref{fig:LIT_pipeline}. Note that we expand the definition of contrast sets in \citet{Gardner2020EvaluatingNM}. We not only apply our generated contrast sets for evaluation but also for augmentation. We also no longer restrict that the perturbations necessarily lead to the change of the labels.

LIT contains seven phenomenon-specific transformation rules for modifying the parse results and can be further extended;
LIT also allows the composition of different transformation rules for complicated perturbations involving multiple linguistic phenomena.

\subsection{Parse and Generation}

    LIT utilizes an existing grammar implementation for parsing and generation, namely the English Resource Grammar (ERG, \citealt{copestake-flickinger-2000-open}).
    ERG is a linguistically motivated broad-coverage grammar for English in the Head-Driven Phrase Structure Grammar framework (HPSG, \citealt{pollard_head-driven_1994, Sag2003}
    ) covering 82.6\% of sentences in Wall Street Journal (WSJ) sections in the Penn Treebank~\citep{marcus-etal-1993-building}.
    ERG is processing-neutral, meaning that it is not limited to either parsing or generation, and can handle both with a grammar processor.
    In this work, we use the ACE parser\footnote{\texttt{http://sweaglesw.org/linguistics/ace/}} as the processor for ERG grammar.

\subsection{Transformation}

    The core part and original contributions of our LIT system are the transformation rules; each rule modifies the parse results from ERG and the ACE parser for one linguistic phenomenon.
    An ERG parse result includes an HPSG syntax tree and a semantic representation in Minimal Recursion Semantics (MRS, \citealt{copestake_minimal_2005}).
    An MRS representation consists of a bag of \emph{elementary predicates} (EPs), each with a \emph{handle} for reference, a set of \emph{handle constraints} that specify relations between handles, a \emph{top} indicating the topmost EP, and an \emph{index} variable for the event described by the entire sentence.
    Every variable has a set of features such as tense and numbers indicating the properties of the entities or the events.
    
    In what follows, we illustrate the application of the transformation rule for it-cleft construction applied to the sentence \emph{Alice saw Bob.}; see Appendix~\ref{app:phenomenon} for a full list of rules.

\input{figtabs/ex_itcleft_transformation}

    For each parse result, LIT generates one transformation for each linguistic phenomenon, obtaining a set of simple transformations.
    Each transformation result in this set can also be fed into LIT as a new base for transformation, allowing different rules to be stacked and producing compositions of transformations.
    LIT uses all transformation results to generate surface sentences.

\input{figtabs/tab_label_rules}
\input{figtabs/tab_example_count}

\subsection{Surface Sentence Selection}

    \paragraph{Selection by ERG:}
        One of the advantages of the LIT system is that the grammar backbone ensures the acceptability of the generated data.
        The ACE parser only generates grammatical sentences, according to the ERG.
        Consequently, ill-formed LIT-transformed results are automatically rejected at the generation phase without additional efforts from the users and developers;
        for instance, even though LIT may produce a representation that would correspond to \emph{*Alice may will see Bob.} \footnote{* means ungrammatical}, such a surface string will not be generated since the ERG does not accept it.
        
        In practice, ERG slightly overgenerates and allows certain ungrammatical strings.
        Such cases are likely too rare to affect the overall quality of the dataset and can often be filtered out during post-selection.
        ERG also cannot rule out grammatically well-formed but semantically unnatural sentences, which limits the data quality for certain constructions, especially for passives.  As a sanity check, we had expert annotators evaluate the generated data and found high agreement on the grammaticality of generated data; the full details are in Appendix~\ref{app:agreement}.
    
    \paragraph{Post-Selection by Pretrained Language Models:}
        ERG often permits multiple strings for a single representation since the meaning-to-form mapping is not unique in natural languages.
        To select the candidate sentence for a specific transformation, LIT employs GPT-2 \cite{radford2018improving} to rank multiple surface sentences generated from the same representation and selects the best one according to their perplexity scores.

\subsection{Phenomena Covered}
\label{coverage}

    LIT is capable of perturbing sentences for seven linguistic transformations:
    polar questions, it-clefts, tense and aspect, modality, negation, passives and subject-object swapping.
    Examples for each transformation are shown in Appendix~\ref{app:phenomenon}.
    LIT also allows different transformations to be stacked where possible.
    LIT can be further extended for more linguistic transformations, and any extension to the LIT system would also receive all of the aforementioned benefits from ACE and ERG.

\subsection{Comparison with Other Approaches}

    \paragraph{Flexibility:}
        LIT covers certain simple constructions that can be handled with a template-based approach, for instance, the subject-object swapping in \citet{mccoy-etal-2019-right}.
        LIT is, however, not limited to template-generated examples and is capable of perturbing naturally-occurring instances.
        
    \paragraph{Plausibility:}
        One special property setting LIT apart from other automatic dataset-construction methods is that LIT uses existing linguistic theories resources as its backbone.
        The use of ERG enables LIT to control data plausibility without human annotation from scratch.

    \paragraph{Modularity:}
        LIT consists of multiple modules: parsing and generation, transformation, and post-selection.
        Extending with more transformation rules, updating ERG (which is still under active development), and including other language models for post-selection can all be handled in the system without major modification to other modules, allowing LIT to be reused for different works.

    \paragraph{Model Agnostic:}
        LIT employs traditional linguistic methods for transforming sentences, and the role of language models is limited to selecting the best one from the strings generated by ERG.
        Contrasting to models trained on specific datasets, the ERG grammar behind LIT does not introduce bias from any specific architecture or dataset.
        This increases the utility of contrast sets generated with LIT as they are likely to be used for testing data-driven models.

\subsection{Sentence Coverage}
\label{sec:sentence_coverage}

    LIT successfully transformed 21.0\% of the sentence pairs in MNLI and 19.7\% in SNLI, with at least one transformed result for each sentence in the pair.
    The number of transformed sentence pairs by phenomenon is shown in Table~\ref{tab:aug_stats}.

%% file: figtabs/ex_itcleft_transformation.tex
\begin{itemize}
    \small{%
    \item[(1)] Original parse result:
    }
\end{itemize}

\begin{lrbox}{0}%
\begin{lstlisting}
[ TOP: h0
  INDEX: e2
      [ e SF: prop TENSE: past ... ]
  RELS: < [ proper_q LBL: h4 ARG0: x3 ... ]
          [ named LBL: h7 ARG0: x3 CARG: "Alice" ]
          [ _see_v_1 LBL: h1 ARG0: e2 ARG1: x3 ARG2: x9 ... ]
          [ proper_q LBL: h10 ARG0: x9 ... ]
          [ named LBL: h13 ARG0: x9 CARG: "Bob" ] >
  HCONS: < h0 qeq h1 h5 qeq h7 h11 qeq h13 > ]
\end{lstlisting}
\end{lrbox}
\scalebox{0.5}{\usebox0}

\resizebox{\linewidth}{!}{%
    \begin{tabular}{l}
        \textbf{TOP:} label of topmost EP \\
        \textbf{INDEX:} the variable associated with the sentential event \\
        \textbf{RELS:} bag of EPs \\
        \textbf{LBL:} label variable for the EP \\
        \textbf{HCONS:} constrains between labels for EPs; $qeq$ denotes a scoping relation \\
    \end{tabular}%
}

\begin{itemize}
    \small{%
    \item[(2)] Inserting it-cleft EP
    }
\end{itemize}

\begin{lrbox}{1}%
\begin{lstlisting}
[ _be_v_itcleft LBL: h14 ARG0: e15 ARG1: x3 ARG2: h1 ]
\end{lstlisting}
\end{lrbox}
\scalebox{0.5}{\usebox1}

\begin{itemize}
    \small{%
    \item[(3)] Connecting top handle $h0$ to it-cleft EP handle (LBL)
    }
\end{itemize}

\begin{lrbox}{2}%
\begin{lstlisting}
HCONS: < h0 qeq h14 h5 qeq h7 h11 qeq h13 > ]
\end{lstlisting}
\end{lrbox}
\scalebox{0.5}{\usebox2}

\begin{itemize}
    \small{%
    \item[(4)] Change sentential semantic index to it-cleft EP's ARG0
    }
\end{itemize}

\begin{lrbox}{3}%
\begin{lstlisting}
INDEX: e15
  [ e SF: prop TENSE: pres ... ]
\end{lstlisting}
\end{lrbox}
\scalebox{0.5}{\usebox3}

\begin{itemize}
    \small{%
    \item[(5)] Final result
    }
\end{itemize}

\begin{lrbox}{4}%
\begin{lstlisting}
[ TOP: h0
  INDEX: e15
    [ e SF: prop TENSE: pres MOOD: indicative PROG: - PERF: - ]
  RELS: < [ proper_q LBL: h4 ... ]
          [ named LBL: h7 ARG0: x3 CARG: "Alice" ]
          [ _see_v_1 LBL: h1 ARG0: e2 ARG1: x3 ARG2: x9 ]
          [ proper_q LBL: h10 ARG0: x9 ... ]
          [ named LBL: h13 ARG0: x9 CARG: "Bob" ]
          [ _be_v_itcleft LBL: h14
            ARG0: e15 ARG1: x3 ARG2: h1 ] >
  HCONS: < h0 qeq h14 h5 qeq h7 h11 qeq h13 > ]
\end{lstlisting}
\end{lrbox}
\scalebox{0.5}{\usebox4}

\medskip

%% file: figtabs/tab_label_rules.tex
\begin{table*}
    \centering
    \resizebox{\textwidth}{!}{%
        \begin{tabular}{llll}
            \toprule
            Transformation
                        & Label         & Sentence 1                                & Sentence 2 \\
            \midrule
            \abb{o;o}    & Contradiction & Alice is driving a car.                   & Alice is playing piano. \\
            \hline
            \abb{i;i}         & Unchanged     & It is Alice who is driving a car.         & It is Alice who is playing piano. \\
            \abb{pa;pa}         & Unchanged     & A car is being driven by Alice.           & Piano is being played by Alice. \\
            \abb{f;p}         & Neutral       & Alice will be driving a car.              & Alice was playing piano. \\
            \abb{m;o}         & Neutral       & Alice may be driving a car.               & Alice is playing piano. \\
            \abb{f;p +i}     & Neutral       & It is Alice who will be driving a car.    & It is Alice who was playing piano. \\
            \abb{f;p +pa}     & Neutral       & A car will be driven by Alice.            & Piano will be played by Alice. \\
            \bottomrule
        \end{tabular}%
    }
    \caption{Examples for label rules used for determining labels of generated data for different transformations. We use \abb{a;b + c} to denote compositional transformation where the premise is transformed with both \abb{a} and \abb{b}, whereas the hypothesis is transformed with \abb{b} and \abb{c}.}
    \label{tab:label rules}
\end{table*}

%% file: figtabs/tab_example_count.tex
\begin{table}[ht!]
    \centering 
    \begin{tabular}{llll|lll}
        \toprule
        & \multicolumn{3}{l}{\# MNLI ex.} & \multicolumn{2}{l}{\# SNLI ex.}\\
        \midrule
         &  \textbf{train} & \textbf{m.} & \textbf{mm.}  &  \textbf{train}  & \textbf{dev}\\
         \midrule
        \abb{o;o} &  392k & 10k & 10k  & 550k  & 10k   \\
        \abb{i;i}  &  13k &  1k & 1k  & 65k  & 1k  \\
        \abb{pa;pa}   &  3k &  236  & 353  & 16k  &  586  \\
        \abb{f;p}  &  3k &  221  & 208  & 6k  &  111  \\
        \abb{p;f} &  3k & 262 & 260 & 7k  &  142   \\
        \abb{m;o}  &  13k& 1k & 1k & 48k & 905\\
        \abb{p;f +i}  &  4k& 288 & 303  & 7k & 122  \\
        \abb{p;f +pa}  & 719& 61 & 82  & 1k & 45  \\
        \abb{f;p +i}  &  4k& 259 & 270  & 6k & 91   \\
        \abb{f;p +pa}  &  727& 59 & 72  & 1k & 37  \\
        \bottomrule
    \end{tabular}
    \caption{Number of examples (ex.) of different transformation rules in MNLI and SNLI parse results. For MNLI, we report example count in training (\textbf{train}) set and development (\textbf{dev}) set. For SNLI, we report example count in training (\textbf{train}) set, matched development  (\textbf{m.}) set, and mismatched development  (\textbf{mm.}) set. }
    \label{tab:aug_stats}
\end{table}

%% file: sec_experiment.tex
Using LIT, we evaluate whether large pretrained models `understand' certain linguistic phenomena through testing them on transformed SNLI and MNLI instances. Specifically, we investigate whether BERT and RoBERTa can successfully predict transformed instances on modality (may), tenses (past; future), passivization, it cleft, and their compositions correctly and consistently. In the following section, we first discuss how we set up our tasks, and then we present our results on simple transformations and composed transformations, respectively.

\subsection{Setup}

\label{applying-lit}

For the purpose of this paper, we formulate our experiment settings as follows. Specifically,  each instance in SNLI/MNLI consists of a hypothesis (e.g., \emph{Some men are playing a sport.}), a premise (e.g. \emph{A soccer game with multiple males playing.}) and their corresponding relationship label (entailment). A dual transformed instance is obtained by applying LIT to either the hypothesis or premise, which may or may not change the label of their relationship (i.e., entailment, neutral, and contradiction). 

\subsubsection{Transforming NLI Datasets with LIT}

 While LIT does not produce laebls after transformation, we apply two  label-changing, two label-preserving, and the relevant compositional transformations listed in Table~\ref{tab:label rules}, with one example per transformation.
 Note that \abb{o;o} means we do not modify the instance.

\input{figtabs/tab_roberta-large_detail_results}

\begin{itemize}
    \item \textbf{Modality} is used to talk about possibilities and necessities beyond what is actually true and is central to natural language semantics \cite{Kratzer1991}. We investigate models' ability to understand the uncertainty expressed in the text by adding `may' to the instance. Thus, a `contradiction' or `entailment' relationship label is changed to `neutral' logically. Specifically, we consider adding `may' to the premise (\abb{m;o}). Note that one can also add `may' to the hypothesis, which we leave for future work.
    
    \item \textbf{Tenses} are used to evaluate sentences at times other than the time of utterance. To probe whether models are able to perform temporal reasoning, we transformed the instances by assigning past tense to hypothesis and future tense to premise (\abb{p;f}) or vice versa (\abb{f;p}), which changes the `contradiction' and `entailment' label to `neutral'.
    
    \item \textbf{Label-preserving Transformations} do not require inferring the label after transformation, which serves to test models' ability to stay consistent with its prediction after some linguistic perturbations. Here, we experiment on passivization (\abb{pa}) and it-cleft (\abb{i}).
    
    \item \textbf{Compositional Transformations} help us further evaluate models' `understanding' of certain linguistic phenomenon. If the models robustly `understand' phenomenon $\alpha$ and $\beta$, composing both should not pose problems to the models. Specifically, we consider adding passivization and it cleft to \abb{p;f} and \abb{f;p} transformations. They are denoted as \abb{p;f +i}, \abb{p;f +p}, \abb{f;p +i}, and \abb{f;p +p} respectively. 
    
\end{itemize}
The statistics for our generated dataset are shown in Table \ref{tab:aug_stats}.
We train two models on two training set.
The original (ORI) training set includes untransformed SNLI training data, whilst the augmented (AUG) training set includes LIT-transformed data with all non-conpositional transformations listed in Table~\ref{tab:aug_stats}.
We test both models' accuracy and consistency for all transformations in the same table.

We use a set of rules to infer the labels of generated pairs (see Table~\ref{tab:label rules}) based on the types of transformation and the original labels.
For instance, originally entailment pairs will turn neutral when `may' is inserted since the `may' modality discharges the truth value of original propositions.
`Decoupling' the tenses of originally present-tense pairs for past/future tense pairs also turns the label to neutral, for events at different times are less likely to affect each other.

We hypothesize that NLI tasks follow logic rules completely and our following experiments also conform to that hypothesis, which legitimize our label-inferring rules.
However exceptions to such rules may occur: \emph{Alice died} nevertheless contradicts \emph{Alice will be eating}, since dying is an event preventing future action of its agent.
Annotation by three experts of 100 randomly chosen transformed pairs shows that 79\% human agreement with the inferred label, with 92\% for label-preserving transformations and 76\% for label-changing transformations.  Future work will explore refinements of our label-assignment procedure.

\input{figtabs/tab_overall_results}
\subsubsection{Probing Models} 
For pretrained language models, we use models from HuggingFace \cite{Wolf2019HuggingFacesTS}.
In this paper, we use {\small\verb|bert-base-uncased|}, {\small\verb|bert-large-uncased|} \cite{Devlin2019BERTPO}, {\small\verb|roberta-base|}, and  {\small\verb|roberta-large|} \cite{Liu2019RoBERTaAR}. For all models, we use Adam to optimize the parameters with an initial learning rate of $5 \times 10^{-5}$. For all the fine-tuning, we use the same seed and train with batch size $32$ for $3$ epochs, the same setting used in \cite{Devlin2019BERTPO}. In this paper, since we never use the development set for early stopping or hyper-parameter tuning (and since MNLI doesn't have a publicly available test set) , we evaluate our models on the development set. Note that MNLI has matched (\textbf{m.}) and mismatched (\textbf{mm.}) test examples, which are derived from the same and different sources as those in the training set, respectively.
\\

\subsubsection{Evaluation Metrics}
To fully evaluate models' performance, we use both accuracy and consistency. While accuracy measures how well a model can accurately predict test instances, consistency measures how robust a model under certain perturbations. We report accuracy on the original test set (Acc@Ori), accuracy on the generated contrast set (Acc@Ctr), and the consistency score (defined below). Note that test sets for different phenomena might be different since we only choose the test instances to be included for each phenomenon if LIT produces contrast instances corresponding to the phenomenon. 

\noindent\textbf{Consistency} In addition to using accuracy to measure models' performance, recent research pays attention to consistency, which provides another perspective to probe models' competence in the real world \citep{trichelair2018reasonable, zhou2019evaluating, Gardner2020EvaluatingNM}. If a model is robust for the given task, then its performance on original and transformed data should be consistent.
For instance, a human is expected to be consistent over the understanding of both a simple sentence and its it-cleft counterpart.
We thus measure consistency by comparing the model's prediction on original and transformed data.
We define consistency for a dual test instance as the match between labels assigned on original and transformed data instances.
Specifically, we define the model to be consistent if a model makes the same label prediction (whether correct or not) for a dual test instance as for the original, and inconsistent otherwise.\footnote{Note that this contrasts with what \citet{Gardner2020EvaluatingNM} call \textit{contrast consistency}, where both predictions additionally have to be both correct.}
We evaluate the model consistency for each type of linguistic transformation to investigate the models' robustness to different linguistic phenomena,
and to examine the differences between the difficulties of different linguistic structures for the models. 

\subsection{Simple Transformations}
\label{simple-transformations}
By perturbing the test instances with our predefined transformations, we aim to probe pre-trained language models' relevant linguistic knowledge and robustness towards those transformations. 

As shown in Table~\ref{tab:mnli_RobL}, RoBERTa, trained on ORI of MNLI, performs worse on contrast sets, especially for label-changing transformations. Label-preserving transformations do not hurt models' performance as much as label-changing transformations. We observed similar trends for other models (see Appendix~\ref{app:full results}. This observation is aligned with \citep{mccoy-etal-2019-right}), which suggests that models are relying on lexical overlaps to infer the relationship between premise and hypothesis. 

Another observation is that RoBERTa does not achieve high consistency in any of the simple transformations.  The poor and inconsistent performance of RoBERTa on our contrast sets shows that even though the model can perform very well on the in-distribution test set, there is still a systematic gap for future models to overcome.

\subsection{Applying LIT for Data Augmentation}
Having shown that pre-trained language models do not generalize well to our generated contrast sets, we ask whether we can `teach' models to recognize those phenomena and make correct predictions accordingly. 

We do this by fine-tuning models on the augmented training data together with the original data. As shown in Table~\ref{tab:aug_overall_results}, we observe that, when training on the augmented data, models preserve their performance on the original test set while improving significantly on the out-of-distribution test sets.

Taking a closer look over the specific phenomenon in Table~\ref{tab:mnli_RobL}, models' performance increases significantly on label-changing contrast sets. This indicates that models improve in terms of `understanding' the role of modality (may) and tenses in natural language inference. Arguably, models may simply memorize the `trick' that modality (may) and tenses (past to future) are associated with label `neutral.' However, we successfully show that we could enable models to learn those `tricks' through data augmentation.  Future work will probe whether models fine-tuned on our augmented data are relying on such heuristics.

The models' performance also increases slightly for label-preserving transformations. However, their consistency does not increase for every transformation, which suggests that data augmentation alone may not suffice for building robust models.

\subsection{Compositional Transformations}
\label{compositional-transformations}

We further investigate the models' performance when multiple transformation rules are composed together and applied to a single sentence. We probe models fine-tuned on the original dataset and the dataset augmented with only simple transformations with our compositional test sets.  If a model learns the linguistic phenomenon systematically, it should perform well on these compositional transformations even without training.  This resembles the zero-shot tests on tasks like SCAN \citep{Lake2018GeneralizationWS}, but applied to naturally occurring linguistic data.\footnote{See \citet{andreas-2020-good} for a  complementary, heuristic-driven approach to compositional data augmentation.}

The bottom-right quadrant of Table~\ref{tab:mnli_RobL} shows that RoBERTa performs very well on compositional transformations when it is fine-tuned only on simple transformations, in some cases (\abb{p;f + pa}) even performing better than on the simple transformation data. Again, we observed similar results across all models (see Appendix~\ref{app:full results}). This suggests that it has learned something systematic about the transformations in the augmented dataset.  

For both \abb{p;f} and \abb{f;p}, RoBERTa performs worse when additionally composing with it-clefts than with passivization.  This suggests that there are differences in the level of systematicity learned for the different transformations, a phenomenon which future work will investigate in more detail.

%% file: figtabs/tab_roberta-large_detail_results.tex
\begin{table*}[ht!]
    \centering 
    \resizebox{\textwidth}{!}{%
        \begin{tabular}{llllllllllll}
            \toprule
            &  & \abb{f;p}  &  \abb{p;f}  & \abb{i;i} & \abb{pa;pa}  & \abb{m;o} & \abb{p;f +i} & \abb{p;f +pa} & \abb{f;p +i} & \abb{f;p +pa}  \\
            \midrule
            \multirow{3}{*}{ORI} 
            & Acc@Ori & 93.21  & 91.60  & 91.86 & 95.28  & 90.90 &  90.97 & 93.44 & 94.21 & 94.92  \\
            & Acc@Ctr & 5.43  & 41.98  & 85.17 & 90.99  & 15.13  &  34.38 & 32.79 & 6.18 &  6.78   \\
            & Consistency &  4.98  & 34.35  & 91.04 & 93.99 & 10.19 &  28.82 & 29.51 & 5.79 & 5.08  \\
            \hline
            &  & \abb{f;p}  &  \abb{p;f}  & \abb{i;i} & \abb{pa;pa}  & \abb{m;o} &  \abb{p;f +i} & \abb{p;f +pa} & \abb{f;p +i} & \abb{f;p +pa}  \\
            \midrule
            \multirow{3}{*}{AUG} 
            & Acc@Ori  &  93.67  &  93.13  & 91.86  &  94.85  &  92.19  &  90.97  & 93.44  & 94.98  &  94.92  \\
            & Acc@Ctr &  99.10  & 99.62  & 89.80  &  91.85  &  99.11  &   87.50  & 98.36  &  76.06  & 94.92   \\
            & Consistency &  92.76  & 92.75  &  95.06  &  94.42  &  91.49  &  78.47  & 91.80  & 71.04  &  89.83  \\
            \bottomrule
        \end{tabular}%
    }
    \caption{Consistency and accuracies of \texttt{roberta-large} over different linguistic phenomena in MNLI. We first train two model separately on the original (ORI) training set and augmented (AUG) training set. Then, we evaluate the trained models on matched genre (\textbf{m.}) for each phenomena. In this table, we report accuracy on the original sentence pair (Acc@Ori), accuracy on the transformed sentence pair (Acc@Ctr), and the model's consistency.}
    \label{tab:mnli_RobL}
\end{table*}

%% file: figtabs/tab_overall_results.tex
\begin{table*}[ht!]
    \centering 
    \begin{tabular}{llllll}
        \toprule
        &  & MNLI &  aug-MNLI &  SNLI &  aug-SNLI  \\
        \hline
        \multirow{3}{*}{ORI} 
            & \texttt{bert-base-uncased}  & 84.31/84.79  & 69.47/69.05  &  90.97   & 46.96  \\
        & \texttt{bert-large-uncased}  & 86.54/86.46 & 71.28/70.34 &  91.78  &  47.72 \\
        & \texttt{roberta-base} &  88.00/87.60 & 71.95/70.58  &  91.86  &  \textbf{47.72}  \\
        & \texttt{roberta-large}  &  \textbf{90.01/90.34} & \textbf{73.78/73.04} &  \textbf{92.83}  &  46.34  \\
        \hline
        \multirow{3}{*}{AUG} 
        & \texttt{bert-base-uncased}  & 84.62/84.45 & 86.60/85.73 &   90.86  &  94.34 \\
        & \texttt{bert-large-uncased}  &  86.24/86.37  &  88.14/87.98 &    91.49  & \textbf{96.00} \\
        & \texttt{roberta-base}  & 87.51/87.52 & 89.66/89.45 &  92.13  &  95.05 \\
        & \texttt{roberta-large}   & \textbf{90.14/89.84} & \textbf{91.47/91.04} &  \textbf{92.53}   & 95.93 \\
        \bottomrule
    \end{tabular}
    \caption{Accuracy on MNLI and SNLI datasets. MNLI results have the format (\textbf{m.}/\textbf{mm.}). SNLI results are on SNLI \textbf{dev}.}
    \label{tab:aug_overall_results}
\end{table*}

%% file: sec_discussion_and_analysis.tex
\label{discussion}

With LIT, we reveal that current high-performance NLI models still suffer from understanding simple linguistic phenomena.  They can be trained to understand these phenomena in a way that appears systematic.  In the remainder, we discuss the limitations of LIT, applying LIT to investigate the systematic deficiency of current large-scale datasets, and potential applications of LIT to tasks other than NLI.

\subsection{Limitations of LIT}

    One major limitation of LIT is the dependency on ERG, which took more than twenty years of human labor and is specifically for English.
    It is possible to swap ERG/ACE parser with data-driven parsers and generators trained on semantic graphbanks, including the DeepBank \citep{flickinger2012deepbank} which uses the same representation frameworks, potentially extending the method to other languages where a broad-coverage hand-crafted grammar is unavailable.
    Using data-driven models, however, does re-introduce possible model bias and uncertainty of robustness.
    Nevertheless, once such a resource is available, LIT provides a method of transforming sentences for data augmentation and integrating linguistic knowledge into a data-driven NLP pipeline.
    
    Future work will also involve expanding the phenomena covered by LIT by generating new transformation rules (cf.~\ref{coverage}). One potential extension is the insertion of control and raising verbs:
    \ex.[(6)] Alice voted for Bob.
        \a. Alice seemed to have voted for Bob.
        \b. Alice wished to vote for Bob.
        \b. Alice persuaded Carol to vote for Bob.

LIT also has a limited coverage, successfully transforming about 20\% of the instances in SNLI (see Section~\ref{coverage}).
The limited coverage may introduce bias in the generated dataset; for instance, the ERG grammar is more likely to fail when parsing complicated sentences.
Nevertheless, we provide a proof of conept that the method can be used to augment data and probe for understanding of the linguistic phenomena of interest here; a higher recall grammar will only improve the situation, and can be easily integrated.
    
\subsection{Analysing Sentence Types in Datasets}
\label{subsec:sentence_types}
    
    In addition to constructing contrast sets, we also used LIT to directly analyze the sentence types in the transformable portion of SNLI and MNLI to investigate the effects of data bias on pretrained models probed in our work.
    For MNLI, we found that 46.4\% sentences are in present tense, 32.2\% in past tense and only 2.95\% in future tense;
    7.27\% sentences are passive, 0.580\% have \emph{may} modality and 0.227\% are it-cleft sentences.
    We found no passive/future or future/passive tense pairs.
    The lack of sentences with \emph{may} modality and mismatched tense pairs may account for the low performance for those transformations before fine-tuning on them.
    It-cleft transformation does not change the meaning and labels, which may explain the high performance despite its rarity in the original data.
    Note that LIT can only detect linguistic phenomena in sentences parsable with ERG (see Section~\ref{sec:sentence_coverage}), but such functionality can still provide important insights on datasets and can be further explored in future works.

%% file: sec_conclusion.tex
We propose Linguistically-Informed Transformations (LIT), a general method to generate contrast sets using an existing linguistic resource. We apply LIT to transform NLI datasets and evaluate current state-of-the-art NLI models. We reveal the systematic gap between current NLI models and an ideal NLI model for NLP practice, which comes from the inadequate coverage of the linguistic phenomenon of SNLI and MNLI. We further show that models can be further improved by using LIT to augment the training data.  Furthermore, models fine-tuned on simple transformations perform very well on compositional transformations, suggesting that fine-tuning provides some systematic understanding of these phenomena.

%% file: appendices/tab_rule_list.tex
\begin{table*}[]
    \centering
    \begin{tabular}{m{0.2\linewidth} m{0.3\linewidth} m{0.3\linewidth}}
        \toprule
        Phenomenon  & original sentence & generated sentence    \\ \hline
        \midrule
        Future & Two guards are standing at the exit. & Two guards will stand at the exit. \\\hline
        Future+It-cleft: AGR1 & The boy is making snowballs. &	It is the boy who will be making snowballs.\\\hline
        Future+It-cleft: AGR2 & People don't play sports. &	It is not sports that will be played by people.\\\hline
        Future+Passive: AGR2 & A woman drills rock. & Rock will be drilled by a woman.\\\hline
        It-cleft: ARG1 & A boy is blowing bubbles & It is a boy who is blowing bubbles.\\\hline
        It-cleft: ARG1+Passive: ARG2 & The man isn't wearing a hat & It is not the man that a hat is being worn by.\\\hline
        It-cleft: ARG2 & A woman is performing music. & It is music that is being performed by a woman.\\\hline
        Modality: may & A person is lounging in a pool & A person may be lounging in a pool.\\\hline
        Negation & Five people tend sheep. & Five people don't tend sheep.\\\hline
        Negation+It cleft: ARG1 & The woman is playing guitar. & It is not the woman who is playing guitar.\\\hline
        Negation+It cleft: ARG2 & The man and woman are buying beer. & It is not beer that is being bought by the man and woman.\\\hline
        Negation+Passive: ARG2 & A woman is riding a bike. & A bike is being ridden by no woman.\\\hline
        Passive: ARG2 & Adults are playing soccer. & Soccer is being played by adults.\\\hline
        Past & There is two cats outside. & There were two cats outside.\\\hline
        Past+It cleft: ARG1 & A woman is mopping. & It is a woman who was mopping.\\\hline
        Past+It cleft: ARG2 & A boy is playing sports. & It is sports that was being played by a boy.\\\hline
        Past+Passive: ARG2 & A man is reading a newspaper.& A newspaper was being read by a man.\\\hline
        Present &  The large pothole in the road was due to bad winter weather.  & The large pothole in the road is due to bad winter weather.\\\hline
        Present+It cleft: ARG1 & The road developed a big hole. & It is the road that develops a big hole.\\\hline
        Present+It cleft: ARG2 & A man ate a stick. & It is a stick which is eaten by a man.\\\hline
        Present+Passive: ARG2 & Two girls pick flowers outside. & Flowers are picked by two girls outside.\\\hline
        Swap subj/obj & The people look at the mountain. & The mountain looks at the people.\\\hline
        Swap subj/obj+It cleft: ARG1 & A woman is playing a board game. & It is a board game that is playing a woman.\\\hline
        Swap subj/obj+It cleft: ARG2 & A girl in a pink top spins a ribbon. & It is a girl in a pink top that is spun by a ribbon.\\\hline
        Swap subj/obj+Passive: ARG2 & A grown woman carries a scooter. & A grown woman is carried by a scooter.\\
        \bottomrule
    \end{tabular}
    \caption{Examples for the full list of rules}
    \label{tab:rule_list}
\end{table*}

%% file: appendices/tab_annotation.tex
\begin{table*}[]
    \centering
    \begin{tabular}{lll}
        \toprule
        Phenomenon      &   Major(\%) & Una.(\%)\\
        \midrule
        Future + Passive: ARG2         & 98   & 74  \\
        It cleft: ARG1 + Passive: ARG2       & 90    &  76 \\
        Future + It cleft: ARG2           & 94 &  76 \\
        Past + Passive: ARG2         & 82 &  66 \\
        Future + It cleft: ARG1          & 98  &  94 \\
        Past + It cleft: ARG2      & 92 &  78 \\
        Past + It cleft: ARG1      & 100  &  94 \\
        Past      & 96  &  88 \\
        Present      & 100  &  100 \\
        Future      & 100   &  86 \\
        Modality: may      & 100  &  98 \\
        \bottomrule
    \end{tabular}
    \caption{The annotators' agreement table for phenomena used for training. We show the percentage of grammatical sentences deemed by majority of our annotators --- Major(\%), and the percentage of unanimous agreement --- Una.(\%)}
    \label{tab:annotation}
\end{table*}

%% file: appendices/mnli_tables.tex
\begin{table*}[ ]
    \centering 
    \begin{tabular}{lllllll}
        \toprule
        &  & \abb{f;p}  &   \abb{p;f} &  \abb{i;i}  & \abb{pa;pa}  & \abb{m;o}\\
        \midrule
        \multirow{3}{*}{ORI} 
        & Acc@Ori & 89.14/86.54 & 87.02/86.54 & 88.36/86.01  & 89.27/88.83 & 88.72/87.20  \\
        & Acc@Ctr & 7.69/11.54  & 41.98/33.85 & 83.21/81.87 & 85.41/85.39 & 13.75/12.80  \\
        & Consistency & 7.69/9.62 & 33.59/26.54 & 88.67/89.95 & 88.41/91.98 & 10.19/9.65  \\
        \hline
         &  & \abb{p;f +i}   & \abb{p;f +pa} & \abb{f;p +i}  & \abb{f;p +pa}  &  \\
        \midrule
         & Acc@Ori & 86.81/86.47 & 88.52/85.37 & 91.89/87.41 & 91.53/87.50 &   \\
         & Acc@Ctr & 37.50/28.05 & 31.15/19.51 & 8.49/10.00 & 6.78/5.56 & \\
         & Consistency & 34.72/24.42 & 22.95/19.51 & 7.34/10.74 & 5.08/9.72 &   \\
        \hline
        &  & \abb{f;p}  &   \abb{p;f} &  \abb{i;i}  & \abb{pa;pa}  & \abb{m;o} \\
        \midrule
        \multirow{3}{*}{AUG} 
        & Acc@Ori & 90.05/86.54 & 86.26/86.92 & 88.05/87.19   & 89.70/89.11 & 88.43/87.66  \\
        & Acc@Ctr & 99.55/97.12  & 98.47/98.85 &  87.74/84.83  & 85.84/85.67 & 99.21/97.31   \\
        & Consistency & 89.59/83.65 & 84.73/85.77 &  93.72/93.30  & 87.55/93.70 & 87.64/85.34 \\
        \hline
          &  & \abb{p;f +i}   & \abb{p;f +pa} &  \abb{f;p +i}  & \abb{f;p +pa}  &  \\
        \midrule
         & Acc@Ori & 86.46/88.45 & 88.52/90.24 &  92.28/88.89 & 91.53/88.89 &   \\
         & Acc@Ctr & 76.39/65.35 & 100.00/95.12 & 46.72/38.15 & 89.83/97.22 &  \\
         & Consistency & 67.01/57.76 & 88.52/85.37 & 39.77/34.44 & 81.36/86.11 &   \\
        \bottomrule
    \end{tabular}
    \caption{Consistency and accuracies of \texttt{bert-base-uncased} over different linguistic phenomena in MNLI. We first train two model separately on the original (ORI) training set and augmented (AUG) training set. Then, we evaluate the trained models on \textbf{dev-m.} and \textbf{dev-mm.} for each phenomena. In this table, we report accuracy on the original sentence pair (Acc@Ori), accuracy on the transformed sentence pair (Acc@Ctr), and the model's consistency. Each accuracy/consistency has the format (\textbf{dev-m.}/\textbf{dev-mm.})}
    \label{tab:results}
\end{table*}

\begin{table*}[ ]
    \centering 
    \begin{tabular}{lllllll}
        \toprule
       &  & \abb{f;p}  &   \abb{p;f} &  \abb{i;i} & \abb{pa;pa}  & \abb{m;o} \\
        \midrule
        \multirow{3}{*}{ORI} 
        & Acc@Ori &  94.12/90.87 &  88.55/88.85 &  89.60/89.85 &  93.56/91.98 &  90.21/89.70    \\
        & Acc@Ctr &  8.60/10.10  &  41.22/33.08 &  84.35/85.02 &  87.12/87.11 &  13.75/12.52  \\
        & Consistency &  7.24/8.65 &  35.88/24.23 &  89.80/91.23 &  88.41/92.84 &  10.68/9.83  \\
        \hline
         &  & \abb{p;f +i}   & \abb{p;f +pa} & \abb{f;p +i}  & \abb{f;p +pa}  &  \\
        \midrule
         & Acc@Ori &  88.89/87.46 &  93.44/90.24 &  93.05/92.96 &  91.53/95.83 &   \\
         & Acc@Ctr &  38.54/27.72 &  32.79/14.63 & 6.95/9.26 &  8.47/5.56 & \\
         & Consistency &  32.99/23.10 &  29.51/12.20 &  5.41/8.15 &  3.39/6.94 &   \\
        \hline
        &  & \abb{f;p}  &   \abb{p;f} &  \abb{i;i} & \abb{pa;pa}  & \abb{m;o} \\
        \midrule
        \multirow{3}{*}{AUG} 
        & Acc@Ori &  90.50/92.31  & 89.31/88.46 & 88.36/89.26  & 91.85/90.54 & 88.63/88.87  \\
        & Acc@Ctr &  100.00/99.04  & 98.85/99.23 & 87.95/88.87 & 88.84/85.96 & 99.31/98.70  \\
        & Consistency &  90.50/91.35  & 88.17/87.69 & 95.06/93.50  & 93.56/90.83 & 87.93/87.76 \\
        \hline
         &  & \abb{p;f +i}   & \abb{p;f +pa} & \abb{f;p +i}  & \abb{f;p +pa}  &  \\
        \midrule
         & Acc@Ori & 88.19/87.46 & 90.16/89.02 &  90.73/91.48 & 91.53/93.06 &   \\
         & Acc@Ctr & 79.51/70.63 & 100.00/96.34 & 59.46/59.26 & 96.61/93.06 &  \\
         & Consistency & 68.40/59.41 & 90.16/85.3791.48 & 52.51/56.67 & 88.14/88.89 &   \\
        \bottomrule
    \end{tabular}
    \caption{Consistency and accuracies of \texttt{bert-large-uncased} over different linguistic phenomena in MNLI. We first train two model separately on the original (ORI) training set and augmented (AUG) training set. Then, we evaluate the trained models on \textbf{dev-m.} and \textbf{dev-mm.} for each phenomena. In this table, we report accuracy on the original sentence pair (Acc@Ori), accuracy on the transformed sentence pair (Acc@Ctr), and the model's consistency. Each accuracy/consistency has the format (\textbf{dev-m.}/\textbf{dev-mm.})}
    \label{tab:results}
\end{table*}

\begin{table*}[ ]
    \centering 
    \begin{tabular}{lllllll}
        \toprule
        &  & \abb{f;p}  &   \abb{p;f} &  \abb{i;i}& \abb{pa;pa}  & \abb{m;o} \\
        \midrule
        \multirow{3}{*}{ORI} 
        & Acc@Ori &  91.40/91.83 &  90.08/90.38 &  90.83/91.23 &  94.42/91.98 &  91.30/91.93   \\
        & Acc@Ctr &  8.14/9.13  &  35.50/23.46 &  85.58/83.84 &  88.84/87.11 &  13.06/11.69 \\
        & Consistency &  10.41/7.69 &  30.15/18.46 &  90.42/89.26 &  91.85/89.40 &  8.11/9.18  \\
        \hline
         &  & \abb{p;f +i}   & \abb{p;f +pa} &  \abb{f;p +i}  & \abb{f;p +pa}  &  \\
        \midrule
         & Acc@Ori &  91.32/92.41 &  93.44/93.90 &  92.66/92.59 &  94.92/95.83 &   \\
         & Acc@Ctr &  29.51/20.79 &  27.87/13.41 &  7.34/7.04 &  8.47/4.17 & \\
         & Consistency &  23.61/18.48 &  21.31/12.20 &  5.41/8.52 &  3.39/5.56 &   \\
        \hline
        &  & \abb{f;p}  &   \abb{p;f} &  \abb{i;i} & \abb{pa;pa}  & \abb{m;o} \\
        \midrule
        \multirow{3}{*}{AUG} 
        & Acc@Ori & 91.40/91.35 & 87.40/93.08 & 89.39/90.94  & 92.70/92.55 & 90.31/91.65  \\
        & Acc@Ctr & 99.10/99.04  & 98.85/98.46 &  87.33/88.87 & 90.13/87.97 & 99.41/97.96  \\
        & Consistency & 91.40/90.38 & 86.26/91.54 &  94.44/95.76 & 92.27/91.40 & 89.71/89.61  \\
        \hline
         &  & \abb{p;f +i}   & \abb{p;f +pa} &  \abb{f;p +i}  & \abb{f;p +pa}  &  \\
        \midrule
         & Acc@Ori & 89.58/92.74 & 91.80/97.56 & 91.89/91.85 & 93.22/93.06 &   \\
         & Acc@Ctr & 91.67/86.47 & 100.00/95.12 & 83.01/78.89 & 94.92/94.44 &  \\
         & Consistency & 82.64/81.19 & 91.80/92.68 & 75.68/72.22 & 88.14/87.50 &   \\
        \bottomrule
    \end{tabular}
    \caption{Consistency and accuracies of \texttt{roberta-base} over different linguistic phenomena in MNLI. We first train two model separately on the original (ORI) training set and augmented (AUG) training set. Then, we evaluate the trained models on \textbf{dev-m.} and \textbf{dev-mm.} for each phenomena. In this table, we report accuracy on the original sentence pair (Acc@Ori), accuracy on the transformed sentence pair (Acc@Ctr), and the model's consistency. Each accuracy/consistency has the format (\textbf{dev-m.}/\textbf{dev-mm.})}
    \label{tab:results}
\end{table*}

\begin{table*}[ ]
    \centering 
    \begin{tabular}{lllllll}
        \toprule
        &  & \abb{f;p}  &   \abb{p;f} &  \abb{i;i} & \abb{pa;pa}  & \abb{m;o} \\
        \midrule
        \multirow{3}{*}{ORI} 
        & Acc@Ori &  93.21/93.75  &  91.60/92.31 &  91.86/92.12  &  95.28/94.84 &  90.90/93.41  \\
        & Acc@Ctr &  5.43/7.69  &  41.98/30.38 &  85.17/85.91  &  90.99/89.40 &  15.13/12.43\\
        & Consistency &  4.98/4.33  &  34.35/24.23 &  91.04/89.46  &  93.99/92.26 &  10.19/9.37 \\
        \hline
         &  & \abb{p;f +i}   & \abb{p;f +pa} &  \abb{f;p +i}  & \abb{f;p +pa}  &  \\
        \midrule
         & Acc@Ori &  90.97/92.74 &  93.44/93.90 &  94.21/95.93 &  94.92/98.61  &   \\
         & Acc@Ctr &  34.38/27.39 &  32.79/23.17 & 6.18/8.89 &  6.78/4.17  & \\
         & Consistency &  28.82/23.43 &  29.51/21.95 &  5.79/9.26 &  5.08/5.56  &   \\
         \hline
        &  & \abb{f;p}  &   \abb{p;f} &  \abb{i;i}  & \abb{pa;pa}  & \abb{m;o} \\
        \midrule
        \multirow{3}{*}{AUG} 
        & Acc@Ori & 93.67/95.19 & 93.13/90.77 & 91.86/92.71 & 94.85/94.56 & 92.19/93.97   \\
        & Acc@Ctr & 99.10/98.08  & 99.62/98.85 & 89.80/90.54 & 91.85/91.12 & 99.11/98.52  \\
        & Consistency & 92.76/93.27 & 92.75/89.62 & 95.06/94.68 & 94.42/92.55 & 91.49/92.49 \\
        \hline
         &  & \abb{p;f +i}   & \abb{p;f +pa} & \abb{f;p +i}  & \abb{f;p +pa}  &  \\
        \midrule
         & Acc@Ori & 90.97/92.08 & 93.44/92.68 & 94.98/97.41 & 94.92/98.61 &   \\
         & Acc@Ctr & 87.50/82.51 & 98.36/97.56 & 76.06/71.11 & 94.92/94.44 &  \\
         & Consistency & 78.47/77.23 & 91.80/90.24 & 71.04/69.26 & 89.83/93.06 &   \\
        \bottomrule
    \end{tabular}
    \caption{Consistency and accuracies of \texttt{roberta-large} over different linguistic phenomena in MNLI. We first train two model separately on the original (ORI) training set and augmented (AUG) training set. Then, we evaluate the trained models on \textbf{dev-m.} and \textbf{dev-mm.} for each phenomena. In this table, we report accuracy on the original sentence pair (Acc@Ori), accuracy on the transformed sentence pair (Acc@Ctr), and the model's consistency. Each accuracy/consistency has the format (\textbf{dev-m.}/\textbf{dev-mm.})}
    \label{tab:results}
\end{table*}

%% file: appendices/snli_tables.tex
\begin{table*}[ht!]
    \centering 
    \begin{tabular}{lllllll}
        \toprule
        &  & \abb{f;p}  &   \abb{p;f} &  \abb{i;i} & \abb{pa;pa}  & \abb{m;o}\\
        \midrule
        \multirow{3}{*}{ORI} 
        & Acc@Ori &  94.59   &  91.55  &  91.33 &  93.09  &  92.49  \\
        & Acc@Ctr &  5.41  &  32.39  &  90.94  &  89.46  &  8.62 \\
        & Consistency &  5.41   &  25.35  &  95.70  &  93.96  &  4.42 \\
        \hline
         &  & \abb{p;f +i}   & \abb{p;f +pa} & \abb{f;p +i}  & \abb{f;p +pa}  &  \\
        \midrule
         & Acc@Ori &  90.16  &  93.33  &  95.60  &  97.30   &   \\
         & Acc@Ctr &  48.36  &  20.00  &  4.40  &   2.70  &  \\
         & Consistency &  40.16  &  17.78  &  4.40  &  5.41   &   \\
        \hline
       &  & \abb{f;p}  &   \abb{p;f} &  \abb{i;i} & \abb{pa;pa}  & \abb{m;o} \\
        \midrule
        \multirow{3}{*}{AUG} 
        & Acc@Ori &  92.79  &  90.85  &   91.56  &  92.57  &  93.26  \\
        & Acc@Ctr &  100.00  &  99.30  &   90.94  &  90.50  &  99.89   \\
        & Consistency &  92.79  &  90.14  &  98.12  &  94.82  &  93.15  \\
        \hline
        &  & \abb{p;f +i}   & \abb{p;f +pa} & \abb{f;p +i}  & \abb{f;p +pa}  &  \\
        \midrule
         & Acc@Ori &  90.98  & 91.11  &  94.51  &  91.89  &   \\
         & Acc@Ctr &  81.97  &  100.00  & 52.75  & 100.00  &  \\
         & Consistency & 74.59  &  91.11  &  47.25  &  91.89  &   \\
        \bottomrule
    \end{tabular}
    \caption{Consistency and accuracies of \texttt{bert-base-uncased} over different linguistic phenomena in SNLI. We first train two model separately on the original (ORI) training set and augmented (AUG) training set. Then, we evaluate the trained models on SNLI development set for each phenomena. In this table, we report accuracy on the original sentence pair (Acc@Ori), accuracy on the transformed sentence pair (Acc@Ctr), and the model's consistency.}
    \label{tab:results}
\end{table*}

\begin{table*}[ht!]
    \centering 
    \begin{tabular}{lllllll}
        \toprule
        &  & \abb{f;p}  &   \abb{p;f} &  \abb{i;i} & \abb{pa;pa}  & \abb{m;o}  \\
        \midrule
        \multirow{3}{*}{ORI} 
        & Acc@Ori &  96.40   &  93.66  &  92.34 &  94.30  &  94.03   \\
        & Acc@Ctr &  7.21  &  47.89  &   91.25  &  89.98  &  6.96 \\
        & Consistency &  5.41   &  42.96  &  96.41  &  91.54  &  3.65   \\
        \hline
          &  & \abb{p;f +i}   & \abb{p;f +pa} & \abb{f;p +i}  & \abb{f;p +pa}  &  \\
        \midrule
         & Acc@Ori &  91.80  &  91.11  &  96.70  &  97.30   &   \\
         & Acc@Ctr &  55.74  &  26.67  &   7.69  &  2.70   & \\
         & Consistency &  50.82  &  31.11  &  6.59  &  5.41   &   \\
         \hline
       &  & \abb{f;p}  &   \abb{p;f} &  \abb{i;i} & \abb{pa;pa}  & \abb{m;o}\\
        \midrule
        \multirow{3}{*}{AUG} 
        & Acc@Ori & 94.59  &  92.25  &  92.58  &  93.78  &  93.48    \\
        & Acc@Ctr &  100.00  &  100.00  &   92.19  &  91.19  &  99.89   \\
        & Consistency &  94.59  &  92.25  &  97.27 &  95.34  &  93.37  \\
        \hline
         &  & \abb{p;f +i}   & \abb{p;f +pa} & \abb{f;p +i}  & \abb{f;p +pa}  &  \\
        \midrule
         & Acc@Ori &  92.62  &  91.11  & 95.60  &  97.30  &   \\
         & Acc@Ctr &  95.08  &  100.00  & 89.01  &  100.00  &  \\
         & Consistency &  87.70  &  91.11  & 84.62  &  97.30  &   \\
        \bottomrule
    \end{tabular}
    \caption{Consistency and accuracies of \texttt{bert-large-uncased} over different linguistic phenomena in SNLI. We first train two model separately on the original (ORI) training set and augmented (AUG) training set. Then, we evaluate the trained models on SNLI development set for each phenomena. In this table, we report accuracy on the original sentence pair (Acc@Ori), accuracy on the transformed sentence pair (Acc@Ctr), and the model's consistency.}
    \label{tab:results}
\end{table*}

\begin{table*}[]
    \centering 
    \begin{tabular}{lllllll}
        \toprule
        &  & \abb{f;p}  &   \abb{p;f} &  \abb{i;i} & \abb{pa;pa}  & \abb{m;o} \\
        \midrule
        \multirow{3}{*}{ORI} 
        & Acc@Ori &  95.50   &  92.96  &  92.81 &  93.78  &  94.59  \\
        & Acc@Ctr &  4.50  &  46.48  &  92.42 &  90.67  &  5.97   \\
        & Consistency &  3.60   &  39.44   &  97.42 &  93.44  &  2.32   \\
        \hline
         &  & \abb{p;f +i}   & \abb{p;f +pa} & \abb{f;p +i}  & \abb{f;p +pa}  &  \\
        \midrule
         & Acc@Ori &  91.80  &  91.11  & 96.70  &  97.30   &   \\
         & Acc@Ctr &  54.92  &  22.22  &  4.40  &  2.70   & \\
         & Consistency &  46.72   &  22.22  & 3.30  &  5.41   &   \\
         \hline
        &  & \abb{f;p}  &   \abb{p;f} &  \abb{i;i} & \abb{pa;pa}  & \abb{m;o} \\
        \midrule
        \multirow{3}{*}{AUG} 
        & Acc@Ori &  96.40  &  93.66  &   92.97 &  93.61  &  94.92  \\
        & Acc@Ctr &  100.00  &  100.00  &   92.50 &  92.06  &  99.89   \\
        & Consistency &  96.40  &  93.66  &  97.81 &  95.34  &  94.81 \\
        \hline
         &  & \abb{p;f +i}   & \abb{p;f +pa} & \abb{f;p +i}  & \abb{f;p +pa}  &  \\
        \midrule
         & Acc@Ori &  92.62  &  91.11  &  96.70  &  97.30  &   \\
         & Acc@Ctr &  78.69  &  100.00  & 57.14  &  100.00  &  \\
         & Consistency &  71.31  &  91.11  & 56.04  &  97.30  &   \\
        \bottomrule
    \end{tabular}
    \caption{Consistency and accuracies of \texttt{roberta-base} over different linguistic phenomena in SNLI. We first train two model separately on the original (ORI) training set and augmented (AUG) training set. Then, we evaluate the trained models on SNLI development set for each phenomena. In this table, we report accuracy on the original sentence pair (Acc@Ori), accuracy on the transformed sentence pair (Acc@Ctr), and the model's consistency.}
    \label{tab:results}
\end{table*}

\begin{table*}[]
    \centering 
    \begin{tabular}{lllllll}
        \toprule
        &  & \abb{f;p}  &   \abb{p;f} &  \abb{i;i} & \abb{pa;pa}  & \abb{m;o} \\
        \midrule
        \multirow{3}{*}{ORI} 
        & Acc@Ori &  97.30   &  95.77  &  93.91 &  94.99  &  95.91  \\
        & Acc@Ctr & 3.60   &  30.28  &  92.03   &  92.23  &  5.41   \\
        & Consistency &  0.90   &  27.46   &   96.56  &  94.47  &  2.65  \\
        \hline
        &  & \abb{p;f +i}   & \abb{p;f +pa} & \abb{f;p +i}  & \abb{f;p +pa}  &  \\
        \midrule
         & Acc@Ori &  95.08  &  95.56  &  96.70  &  94.59   &   \\
         & Acc@Ctr &  45.08  &  37.78  &  4.40  &  0.00   & \\
         & Consistency &  41.80  &  37.78  &  1.10  &  5.41   &   \\
        \hline
       &  & \abb{f;p}  &   \abb{p;f} &  \abb{i;i} & \abb{pa;pa}  & \abb{m;o} \\
        \midrule
        \multirow{3}{*}{AUG} 
        & Acc@Ori &  95.50  &  94.37  &  92.89 &  93.96  &  94.48 \\
        & Acc@Ctr &  100.00  &  100.00  &  92.58 &  93.09  &  99.89  \\
        & Consistency &  95.50  &  94.37  &  98.12 &  97.75  &  94.36   \\
        \hline
         &  & \abb{p;f +i}   & \abb{p;f +pa} & \abb{f;p +i}  & \abb{f;p +pa}  &  \\
        \midrule
         & Acc@Ori &  93.44  &  91.11  &  95.60  &  91.89  &   \\
         & Acc@Ctr &  95.08  &  100.00  & 68.13  &  100.00  &  \\
         & Consistency & 88.52  & 91.11   & 65.93  &  91.89  &   \\
        \bottomrule
    \end{tabular}
    \caption{Consistency and accuracies of \texttt{roberta-large} over different linguistic phenomena in SNLI. We first train two model separately on the original (ORI) training set and augmented (AUG) training set. Then, we evaluate the trained models on SNLI development set for each phenomena. In this table, we report accuracy on the original sentence pair (Acc@Ori), accuracy on the transformed sentence pair (Acc@Ctr), and the model's consistency.}
    \label{tab:results}
\end{table*}

%% file: main.bbl
\begin{thebibliography}{30}
\expandafter\ifx\csname natexlab\endcsname\relax\def\natexlab#1{#1}\fi

\bibitem[{Andreas(2020)}]{andreas-2020-good}
Jacob Andreas. 2020.
\newblock \href {https://doi.org/10.18653/v1/2020.acl-main.676} {Good-enough
  compositional data augmentation}.
\newblock In \emph{Proceedings of the 58th Annual Meeting of the Association
  for Computational Linguistics}, pages 7556--7566, Online. Association for
  Computational Linguistics.

\bibitem[{Bowman et~al.(2015)Bowman, Angeli, Potts, and
  Manning}]{bowman-etal-2015-large}
Samuel~R. Bowman, Gabor Angeli, Christopher Potts, and Christopher~D. Manning.
  2015.
\newblock \href {https://doi.org/10.18653/v1/D15-1075} {A large annotated
  corpus for learning natural language inference}.
\newblock In \emph{Proceedings of the 2015 Conference on Empirical Methods in
  Natural Language Processing}, pages 632--642, Lisbon, Portugal. Association
  for Computational Linguistics.

\bibitem[{Bras et~al.(2020)Bras, Swayamdipta, Bhagavatula, Zellers, Peters,
  Sabharwal, and Choi}]{bras2020adversarial}
Ronan~Le Bras, Swabha Swayamdipta, Chandra Bhagavatula, Rowan Zellers,
  Matthew~E. Peters, Ashish Sabharwal, and Yejin Choi. 2020.
\newblock \href {http://arxiv.org/abs/2002.04108} {Adversarial filters of
  dataset biases}.

\bibitem[{Copestake and Flickinger(2000)}]{copestake-flickinger-2000-open}
Ann Copestake and Dan Flickinger. 2000.
\newblock \href {http://www.lrec-conf.org/proceedings/lrec2000/pdf/371.pdf} {An
  open source grammar development environment and broad-coverage {E}nglish
  grammar using {HPSG}}.
\newblock In \emph{Proceedings of the Second International Conference on
  Language Resources and Evaluation ({LREC}{'}00)}, Athens, Greece. European
  Language Resources Association (ELRA).

\bibitem[{Copestake et~al.(2005)Copestake, Flickinger, Pollard, and
  Sag}]{copestake_minimal_2005}
Ann Copestake, Dan Flickinger, Carl Pollard, and Ivan~A. Sag. 2005.
\newblock \href {https://doi.org/10.1007/s11168-006-6327-9} {Minimal
  {Recursion} {Semantics}: {An} {Introduction}}.
\newblock \emph{Research on Language and Computation}, 3(2-3):281--332.

\bibitem[{Devlin et~al.(2019)Devlin, Chang, Lee, and
  Toutanova}]{Devlin2019BERTPO}
Jacob Devlin, Ming-Wei Chang, Kenton Lee, and Kristina Toutanova. 2019.
\newblock Bert: Pre-training of deep bidirectional transformers for language
  understanding.
\newblock In \emph{NAACL-HLT}.

\bibitem[{Dinan et~al.(2019)Dinan, Humeau, Chintagunta, and
  Weston}]{Dinan_2019}
Emily Dinan, Samuel Humeau, Bharath Chintagunta, and Jason Weston. 2019.
\newblock \href {https://doi.org/10.18653/v1/d19-1461} {Build it break it fix
  it for dialogue safety: Robustness from adversarial human attack}.
\newblock \emph{Proceedings of the 2019 Conference on Empirical Methods in
  Natural Language Processing and the 9th International Joint Conference on
  Natural Language Processing (EMNLP-IJCNLP)}.

\bibitem[{Flickinger et~al.(2012)Flickinger, Zhang, and
  Kordoni}]{flickinger2012deepbank}
Dan Flickinger, Yi~Zhang, and Valia Kordoni. 2012.
\newblock Deepbank. a dynamically annotated treebank of the wall street
  journal.
\newblock In \emph{Proceedings of the 11th International Workshop on Treebanks
  and Linguistic Theories}, pages 85--96.

\bibitem[{Gardner et~al.(2020)Gardner, Artzi, Basmova, Berant, Bogin, Chen,
  Dasigi, Dua, Elazar, Gottumukkala, Gupta, Hajishirzi, Ilharco, Khashabi, Lin,
  Liu, Liu, Mulcaire, Ning, Singh, Smith, Subramanian, Tsarfaty, Wallace,
  Zhang, and Zhou}]{Gardner2020EvaluatingNM}
Matt Gardner, Yoav Artzi, Victoria Basmova, Jonathan Berant, Ben Bogin, Sihao
  Chen, Pradeep Dasigi, Dheeru Dua, Yanai Elazar, Ananth Gottumukkala, Nitish
  Gupta, Hanna Hajishirzi, Gabriel Ilharco, Daniel Khashabi, Kevin Lin,
  Jiangming Liu, Nelson~F. Liu, Phoebe Mulcaire, Qiang Ning, Sameer Singh,
  Noah~A. Smith, Sanjay Subramanian, Reut Tsarfaty, Eric Wallace, Ally~Quan
  Zhang, and Ben Zhou. 2020.
\newblock Evaluating nlp models via contrast sets.
\newblock \emph{ArXiv}, abs/2004.02709.

\bibitem[{Geva et~al.(2019)Geva, Goldberg, and
  Berant}]{geva-etal-2019-modeling}
Mor Geva, Yoav Goldberg, and Jonathan Berant. 2019.
\newblock \href {https://doi.org/10.18653/v1/D19-1107} {Are we modeling the
  task or the annotator? an investigation of annotator bias in natural language
  understanding datasets}.
\newblock In \emph{Proceedings of the 2019 Conference on Empirical Methods in
  Natural Language Processing and the 9th International Joint Conference on
  Natural Language Processing (EMNLP-IJCNLP)}, pages 1161--1166, Hong Kong,
  China. Association for Computational Linguistics.

\bibitem[{Gururangan et~al.(2018)Gururangan, Swayamdipta, Levy, Schwartz,
  Bowman, and Smith}]{gururangan-etal-2018-annotation}
Suchin Gururangan, Swabha Swayamdipta, Omer Levy, Roy Schwartz, Samuel Bowman,
  and Noah~A. Smith. 2018.
\newblock \href {https://doi.org/10.18653/v1/N18-2017} {Annotation artifacts in
  natural language inference data}.
\newblock In \emph{Proceedings of the 2018 Conference of the North {A}merican
  Chapter of the Association for Computational Linguistics: Human Language
  Technologies, Volume 2 (Short Papers)}, pages 107--112, New Orleans,
  Louisiana. Association for Computational Linguistics.

\bibitem[{Howard and Ruder(2018)}]{howard-ruder-2018-universal}
Jeremy Howard and Sebastian Ruder. 2018.
\newblock \href {https://doi.org/10.18653/v1/P18-1031} {Universal language
  model fine-tuning for text classification}.
\newblock In \emph{Proceedings of the 56th Annual Meeting of the Association
  for Computational Linguistics (Volume 1: Long Papers)}, pages 328--339,
  Melbourne, Australia. Association for Computational Linguistics.

\bibitem[{Jin et~al.(2019{\natexlab{a}})Jin, Jin, Zhou, and
  Szolovits}]{jin2019bert}
Di~Jin, Zhijing Jin, Joey~Tianyi Zhou, and Peter Szolovits. 2019{\natexlab{a}}.
\newblock \href {http://arxiv.org/abs/1907.11932} {Is bert really robust? a
  strong baseline for natural language attack on text classification and
  entailment}.

\bibitem[{Jin et~al.(2019{\natexlab{b}})Jin, Jin, Zhou, and
  Szolovits}]{Jin2019IsBR}
Di~Jin, Zhijing Jin, Joey~Tianyi Zhou, and Peter Szolovits. 2019{\natexlab{b}}.
\newblock Is bert really robust? natural language attack on text classification
  and entailment.
\newblock \emph{ArXiv}, abs/1907.11932.

\bibitem[{Kaushik et~al.(2020)Kaushik, Hovy, and Lipton}]{Kaushik2020Learning}
Divyansh Kaushik, Eduard Hovy, and Zachary Lipton. 2020.
\newblock \href {https://openreview.net/forum?id=Sklgs0NFvr} {Learning the
  difference that makes a difference with counterfactually-augmented data}.
\newblock In \emph{International Conference on Learning Representations}.

\bibitem[{Kratzer(1991)}]{Kratzer1991}
Angelika Kratzer. 1991.
\newblock Modality. in semantics: An international handbook of contemporary
  research.

\bibitem[{Lake and Baroni(2018)}]{Lake2018GeneralizationWS}
Brenden~M. Lake and Marco Baroni. 2018.
\newblock Generalization without systematicity: On the compositional skills of
  sequence-to-sequence recurrent networks.
\newblock In \emph{ICML}.

\bibitem[{Liu et~al.(2019)Liu, Ott, Goyal, Du, Joshi, Chen, Levy, Lewis,
  Zettlemoyer, and Stoyanov}]{Liu2019RoBERTaAR}
Yinhan Liu, Myle Ott, Naman Goyal, Jingfei Du, Mandar Joshi, Danqi Chen, Omer
  Levy, Mike Lewis, Luke Zettlemoyer, and Veselin Stoyanov. 2019.
\newblock Roberta: A robustly optimized bert pretraining approach.
\newblock \emph{ArXiv}, abs/1907.11692.

\bibitem[{Marcus et~al.(1993)Marcus, Santorini, and
  Marcinkiewicz}]{marcus-etal-1993-building}
Mitchell~P. Marcus, Beatrice Santorini, and Mary~Ann Marcinkiewicz. 1993.
\newblock \href {https://www.aclweb.org/anthology/J93-2004} {Building a large
  annotated corpus of {E}nglish: The {P}enn {T}reebank}.
\newblock \emph{Computational Linguistics}, 19(2):313--330.

\bibitem[{McCoy et~al.(2019)McCoy, Pavlick, and Linzen}]{mccoy-etal-2019-right}
Tom McCoy, Ellie Pavlick, and Tal Linzen. 2019.
\newblock \href {https://doi.org/10.18653/v1/P19-1334} {Right for the wrong
  reasons: Diagnosing syntactic heuristics in natural language inference}.
\newblock In \emph{Proceedings of the 57th Annual Meeting of the Association
  for Computational Linguistics}, pages 3428--3448, Florence, Italy.
  Association for Computational Linguistics.

\bibitem[{Min et~al.(2020)Min, McCoy, Das, Pitler, and
  Linzen}]{min2020augmentation}
Junghyun Min, R.~Thomas McCoy, Dipanjan Das, Emily Pitler, and Tal Linzen.
  2020.
\newblock Syntactic data augmentation increases robustness to inference
  heuristics.
\newblock In \emph{Proceedings of the 58th Annual Meeting of the Association
  for Computational Linguistics}, Seattle, Washington. Association for
  Computational Linguistics.

\bibitem[{Peters et~al.(2018)Peters, Neumann, Iyyer, Gardner, Clark, Lee, and
  Zettlemoyer}]{Peters:2018}
Matthew~E. Peters, Mark Neumann, Mohit Iyyer, Matt Gardner, Christopher Clark,
  Kenton Lee, and Luke Zettlemoyer. 2018.
\newblock Deep contextualized word representations.
\newblock In \emph{Proc. of NAACL}.

\bibitem[{Pollard and Sag(1994)}]{pollard_head-driven_1994}
Carl~Jesse Pollard and Ivan~A. Sag. 1994.
\newblock \emph{Head-driven phrase structure grammar}.
\newblock Studies in contemporary linguistics. Center for the Study of Language
  and Information ; University of Chicago Press, Stanford : Chicago.

\bibitem[{Radford et~al.(2018)Radford, Narasimhan, Salimans, and
  Sutskever}]{radford2018improving}
Alec Radford, Karthik Narasimhan, Tim Salimans, and Ilya Sutskever. 2018.
\newblock Improving language understanding by generative pre-training.

\bibitem[{Radford et~al.(2019)Radford, Wu, Child, Luan, Amodei, and
  Sutskever}]{radford2019language}
Alec Radford, Jeff Wu, Rewon Child, David Luan, Dario Amodei, and Ilya
  Sutskever. 2019.
\newblock Language models are unsupervised multitask learners.

\bibitem[{Sag et~al.(2003)Sag, Wasow, and Bender}]{Sag2003}
Ivan~A Sag, Thomas Wasow, and Emily~M Bender. 2003.
\newblock \emph{{Syntactic Theory: A Formal Introduction}}, volume 152 of
  \emph{CSLI Lecture Notes}.
\newblock CSLI Publications.

\bibitem[{Trichelair et~al.(2018)Trichelair, Emami, Trischler, Suleman, and
  Cheung}]{trichelair2018reasonable}
Paul Trichelair, Ali Emami, Adam Trischler, Kaheer Suleman, and Jackie Chi~Kit
  Cheung. 2018.
\newblock \href {http://arxiv.org/abs/1811.01778} {How reasonable are
  common-sense reasoning tasks: A case-study on the winograd schema challenge
  and swag}.

\bibitem[{Williams et~al.(2018)Williams, Nangia, and Bowman}]{N18-1101}
Adina Williams, Nikita Nangia, and Samuel Bowman. 2018.
\newblock \href {http://aclweb.org/anthology/N18-1101} {A broad-coverage
  challenge corpus for sentence understanding through inference}.
\newblock In \emph{Proceedings of the 2018 Conference of the North American
  Chapter of the Association for Computational Linguistics: Human Language
  Technologies, Volume 1 (Long Papers)}, pages 1112--1122. Association for
  Computational Linguistics.

\bibitem[{Wolf et~al.(2019)Wolf, Debut, Sanh, Chaumond, Delangue, Moi, Cistac,
  Rault, Louf, Funtowicz, and Brew}]{Wolf2019HuggingFacesTS}
Thomas Wolf, Lysandre Debut, Victor Sanh, Julien Chaumond, Clement Delangue,
  Anthony Moi, Pierric Cistac, Tim Rault, R'emi Louf, Morgan Funtowicz, and
  Jamie Brew. 2019.
\newblock Huggingface's transformers: State-of-the-art natural language
  processing.
\newblock \emph{ArXiv}, abs/1910.03771.

\bibitem[{Zhou et~al.(2019)Zhou, Zhang, Cui, and Huang}]{zhou2019evaluating}
Xuhui Zhou, Yue Zhang, Leyang Cui, and Dandan Huang. 2019.
\newblock \href {http://arxiv.org/abs/1911.11931} {Evaluating commonsense in
  pre-trained language models}.

\end{thebibliography}
